\documentclass[sigconf]{acmart}

\usepackage{enumitem}
\usepackage{multirow}
\usepackage{amsmath,bm}
\usepackage{mathtools}
\usepackage{flushend}
\usepackage[ruled,vlined]{algorithm2e}
\usepackage[font=normalsize]{subfig}

\newtheorem{definition}{Definition}
\newtheorem{theorem}{Theorem}
\newtheorem{lemma}{Lemma}

\newcommand{\slfrac}[2]{\left.#1\middle/#2\right.}

\DeclareMathOperator*{\argmin}{arg\,min}

\AtBeginDocument{%
  \providecommand\BibTeX{{%
    \normalfont B\kern-0.5em{\scshape i\kern-0.25em b}\kern-0.8em\TeX}}}

\begin{document}

\title{Differentially Private Counterfactuals via Functional Mechanism}

\author{Fan Yang$^1$, Qizhang Feng$^2$, Kaixiong Zhou$^1$, Jiahao Chen$^3$, Xia Hu$^1$}
\affiliation{
  \institution{$^1$Department of Computer Science, Rice University, Houston, TX, USA} 
  \institution{$^2$Department of Computer Science and Engineering, Texas A\&M University, College Station, TX, USA} 
  \institution{$^3$J.P. Morgan AI Research, New York, NY, USA}
  }
 \email{{fyang, kaixiong.zhou, xia.hu}@rice.edu, qf31@tamu.edu, jiahao.chen@jpmchase.com}

\renewcommand{\shortauthors}{F. Yang et al.}

\begin{abstract}
\emph{Counterfactual}, serving as one emerging type of model explanation, has attracted tons of attentions recently from both industry and academia. Different from the conventional feature-based explanations (e.g., \emph{attributions}), counterfactuals are a series of hypothetical samples which can flip model decisions with minimal perturbations on queries. Given valid counterfactuals, humans are capable of reasoning under ``what-if'' circumstances, so as to better understand the model decision boundaries. However, releasing counterfactuals could be detrimental, since it may unintentionally leak sensitive information to adversaries, which brings about higher risks on both \emph{model security} and \emph{data privacy}. To bridge the gap, in this paper, we propose a novel framework to generate differentially private counterfactual (DPC) without touching the deployed model or explanation set, where noises are injected for protection while maintaining the explanation roles of counterfactual. In particular, we train an autoencoder with the functional mechanism to construct noisy class prototypes, and then derive the DPC from the latent prototypes based on the post-processing immunity of differential privacy. Further evaluations demonstrate the effectiveness of the proposed framework, showing that DPC can successfully relieve the risks on both extraction and inference attacks. 
\end{abstract}


\maketitle

\section{Introduction}

The past decade has witnessed a great success of machine learning (ML) models in many fields, covering lots of online services and practical applications~\cite{shinde2018review,sharma2021machine}. For high-stake scenarios, such as policy making~\cite{brennan2013emergence} and financial analysis~\cite{dixon2020machine}, interpreting model behaviors is becoming increasingly important and necessary, since stakeholders may need to better understand the model decisions before any actions in the real world. Various interpretation techniques~\cite{du2019techniques,murdoch2019definitions} have thus been proposed to handle the ML explanability, including feature attribution~\cite{ribeiro2016should,lundberg2017unified,sundararajan2017axiomatic}, influential sample~\cite{kim2016examples,koh2017understanding} and etc.

\emph{Counterfactual}~\cite{wachter2017counterfactual}, emerging as a new form of model explanation, has raised much attentions of researchers and practitioners due to its strong capability of reasoning for humans. Instead of simply showing the factors that contribute most to certain predictions, counterfactuals are able to help understand how particular queries can cross over the decision boundaries to a preferred output with minimal perturbations. Such advantage enables humans to conduct reasoning under the ``what-if'' circumstances, and shows the potential actions to take for altering model decisions. Essentially, counterfactuals are hypothetical data samples synthesized within certain distributions, which may not necessarily exist in the real world. Given a loan rejection case for example, valid counterfactuals would suggest some specific changes on the profile to make it approved, such as increasing annual income from $\$60,000$ to $\$80,000$ or improving education level from High-School to College. 

Nevertheless, releasing model explanations could be risky, since meaningful explanations always contain additional information on how model works. Adversaries may purposely collect those released explanations to infer model properties, which can cause serious issues on both \emph{model security}~\cite{milli2019model} and \emph{data privacy}~\cite{shokri2021privacy}. Counterfactual explanation can make such threat even more significant, considering the fact that it directly reveals the decision boundaries at local points. Studies in work~\cite{aivodji2020model} show that adversaries are able to extract high-fidelity models from the collected counterfactuals with only limited number of queries, which is demonstrated to be a more effective attack than traditional methods. Thus, to avoid the ML model from being maliciously inferred, there is of great importance to investigate how to generate counterfactuals which are both \emph{protective} and \emph{informative} for the explanation process. 

Existing efforts for relieving risky explanations mainly focus on the integration of differential privacy (DP)~\cite{dwork2008differential,dwork2014algorithmic}, where certain noises are manually added into the explanation pipeline to protect the sensitive information from leaking. To effectively inject DP into explanations, there are two major methodologies from previous work. One way~\cite{mochaourab2021robust} is to inject DP into the target ML model by training, and the other way~\cite{patel2020model} is to inject DP through an explanation set labelled by the model. However, both methodologies could be limited when we try to inject DP into counterfactuals. First, injecting DP during the training can result in a decrease on model accuracy and an amplification on model bias~\cite{bagdasaryan2019differential}, which typically lowers the quality of the derived counterfactuals for reasoning~\cite{verma2020counterfactual}. Besides, the explanation set of counterfactual is quite different from that of the feature-based ones (e.g., LIME~\cite{ribeiro2016should}), where each instance indicates a data trajectory consisting of multiple sequential samples for explanation~\cite{wachter2017counterfactual,keane2020good,naumann2021consequence}. Such distinction further makes it hard to directly apply existing DP schemes (e.g., Laplace mechanism~\cite{koufogiannis2015optimality,phan2017adaptive}) to the counterfactual derivation. 

To bridge the gap, in this paper, we propose a novel explanation framework to generate \textbf{\underline{D}}ifferentially \textbf{\underline{P}}rivate \textbf{\underline{C}}ounterfactual (\textbf{DPC}), aiming to relieve the deployed ML model from being maliciously inferred while maintaining the explanation quality of counterfactual. In particular, we first construct the noisy class prototypes over the training set, through an autoencoder specifically trained with the functional mechanism~\cite{zhang2012functional}, where the DP is injected by the perturbed objective function. Based on the obtained prototype representations in latent space, counterfactual samples are then derived with the regularizations to certain queries and data distributions. The overall DP badge of our proposed framework can be claimed by the post-processing immunity~\cite{dwork2014algorithmic,zhu2021bias}, which indicates that a differentially private output can be transformed using arbitrary randomized mappings without impacting its DP guarantees. The merits of the proposed DPC framework mainly lie in two folds. Firstly, our DPC is typically derived in a post-hoc manner, which does not require any additional training for the target model and thus has no influence on the original model performance. Secondly, the proposed DPC only needs data access to the training set, and does not require extra explanation set for applying DP schemes. In the experiments, we evaluate the proposed DPC on several real-world datasets, considering practical attack scenarios where adversaries try to infer model properties with collected counterfactuals. Empirical results further demonstrate the effectiveness of our DPC, validating that it can successfully relieve the risks from model extraction, membership inference, as well as attribute inference. Our major contributions are summarized as follows:

\begin{itemize}[leftmargin=*]
\item Propose a novel explanation framework to derive counterfactuals with DP guarantees, which follows a total post-hoc manner and only requires the data access for training set; 

\item Theoretically prove the DP badge of the proposed framework based on the post-processing immunity, and derive the sensitivity upper bound for adding noises with the functional mechanism; 

\item Empirically evaluate DPC in real-world datasets, and validate its effectiveness in protecting deployed models from extraction and inference attacks while serving the explanation roles. 
\end{itemize}

\section{Preliminaries} 

In this section, we briefly introduce the counterfactual explanation problem, as well as the concept of DP and the employed mechanism. 

\textbf{Counterfactual Explanation.} This is one particular interpretation technique to help humans better understand the model behaviors. Counterfactuals are typically developed from the example-based reasoning~\cite{watson1994case,rissland2012example,richter2016case} framework, which provide hypothetical data samples to show insights on decision boundaries. Consider a simple case for example, where we have a binary classification model $f: \mathbb{R}^{d} \rightarrow \{-1,1\}$ with $-1$ and $1$ respectively denoting the undesired and desired outputs. The counterfactual explanation problem for model $f$ can then be generally formulated as:
\begin{equation}\label{eq_cf}
  \begin{split}
    \mathbf{x}^{*} & = \ \argmin\nolimits_{\mathbf{x}\sim\mathcal{X}} \ l(\mathbf{x}, \mathbf{q}),    \\  
    \mathrm{s.t.} \quad & f(\mathbf{q})=-1; \  f(\mathbf{x}^{*})=1,
  \end{split} 
\end{equation} 
where $\mathbf{q}$ is the query of interest, and $\mathbf{x}^{*}$ represents the derived counterfactual sample for explanation. In Eq.~\ref{eq_cf}, $\mathcal{X}$ indicates a certain data distribution of the observed data space $\mathbb{R}^{d}$, and $l: \mathbb{R}^{d} \times \mathbb{R}^{d} \rightarrow \mathcal{R}^{+}$ denotes a distance measure between two data samples. Based on Eq.~\ref{eq_cf}, it is noted that counterfactual explanation aims to find proper in-distribution samples which can flip the model decision to desired outputs, while minimizing the distance between the query and hypothetical samples. Since the derived $\mathbf{x}^{*}$ is generally close to the decision boundaries, it could be risky when those counterfactuals are maliciously collected to infer model properties. In this paper, we focus on the counterfactual scheme that can be both protective and informative for interpreting model behaviors. 

\textbf{Differential Privacy.} DP is a rigorous mathematical definition of privacy in the context of statistical ML. To ensure the protection, the model mapping process is required to be conducted with an algorithm that satisfies \emph{$\epsilon$-differential privacy}~\cite{dwork2006calibrating}, which is defined over the neighbor sets differing by only one tuple. Specifically, a randomized algorithm $\mathcal{A}$ is said to fulfill $\epsilon$-differential privacy, if and only if the following inequality holds for any output $\mathcal{O}$ and any two neighbor sets $\mathcal{D}_1$, $\mathcal{D}_2$: 
\begin{equation}\label{eq_dp}
  \frac{\mathrm{Pr}\left[\mathcal{A}(\mathcal{D}_1)=\mathcal{O}\right]}{\mathrm{Pr}\left[\mathcal{A}(\mathcal{D}_2)=\mathcal{O}\right]} \leq \mathrm{e}^{\epsilon}, 
\end{equation} 
where $\epsilon$ is usually referred as the privacy budget for protection. Typically, smaller budgets yield stronger privacy guarantees. When $\epsilon \rightarrow 0$ in Eq.~\ref{eq_dp}, the output distribution of $\mathcal{A}$ is roughly the same for any two neighbor sets, which indicates that the output of $\mathcal{A}$ does not leak significant information about particular tuples in original data. The privacy is thus preserved with a differential view on the mapping process. In this paper, we try to inject DP into the derived counterfactuals, so as to effectively prevent the information leakage from model explanations. 

\textbf{Functional Mechanism.} It is one specific strategy to achieve DP in ML models, which ensures privacy by perturbing the objective function for training~\cite{zhang2012functional}. Consider an ML model, trained on dataset $\mathcal{D}$, with parameter set $\mathbf{w}^{*}$ that minimizes an objective $L_{\mathcal{D}}(\mathbf{w})=\sum_{t_{i}\in \mathcal{D}}L(t_{i},\mathbf{w})$. Functional mechanism perturbs $L_{\mathcal{D}}(\mathbf{w})$ to $\widetilde{L}_{\mathcal{D}}(\mathbf{w})$ by injecting Laplace noises into the coefficients of its polynomial representation, which is generally guaranteed by the Stone-Weierstrass theorem in~\cite{rudin1964principles}. Then, the returned parameter set $\widetilde{\mathbf{w}}^{*}$ minimizing the perturbed loss $\widetilde{L}_{\mathcal{D}}(\mathbf{w})$ can be derived by:
\begin{equation}\label{eq_fm}
  \widetilde{\mathbf{w}}^{*} = \argmin_{\mathbf{w}} \widetilde{L}_{\mathcal{D}}(\mathbf{w}). 
\end{equation} 
Within the process in Eq.~\ref{eq_fm}, it is observed that functional mechanism does not add noises directly to $\mathbf{w}^{*}$, which saves the efforts in analyzing the sensitivity of $\mathbf{w}^{*}$ on $\mathcal{D}$. Considering the complex correlation between $\mathcal{D}$ and $\mathbf{w}^{*}$ in some cases (e.g., deep neural networks~\cite{lecun2015deep,he2016deep,vaswani2017attention}), functional mechanism could be much advantageous in achieving DP for those complicated models. In this paper, we utilize the functional mechanism to achieve DP in an autoencoder, which is used to construct class prototypes in the latent space for counterfactual search.

\section{Counterfactual Framework Design} 

In this section, we formally define the concept of the proposed DPC. Then, we introduce the designed framework to effectively derive DPC for interpreting model behaviors.

\begin{figure*}[t] 
\centering 
\includegraphics[width=0.96\textwidth, height=0.339\textwidth]{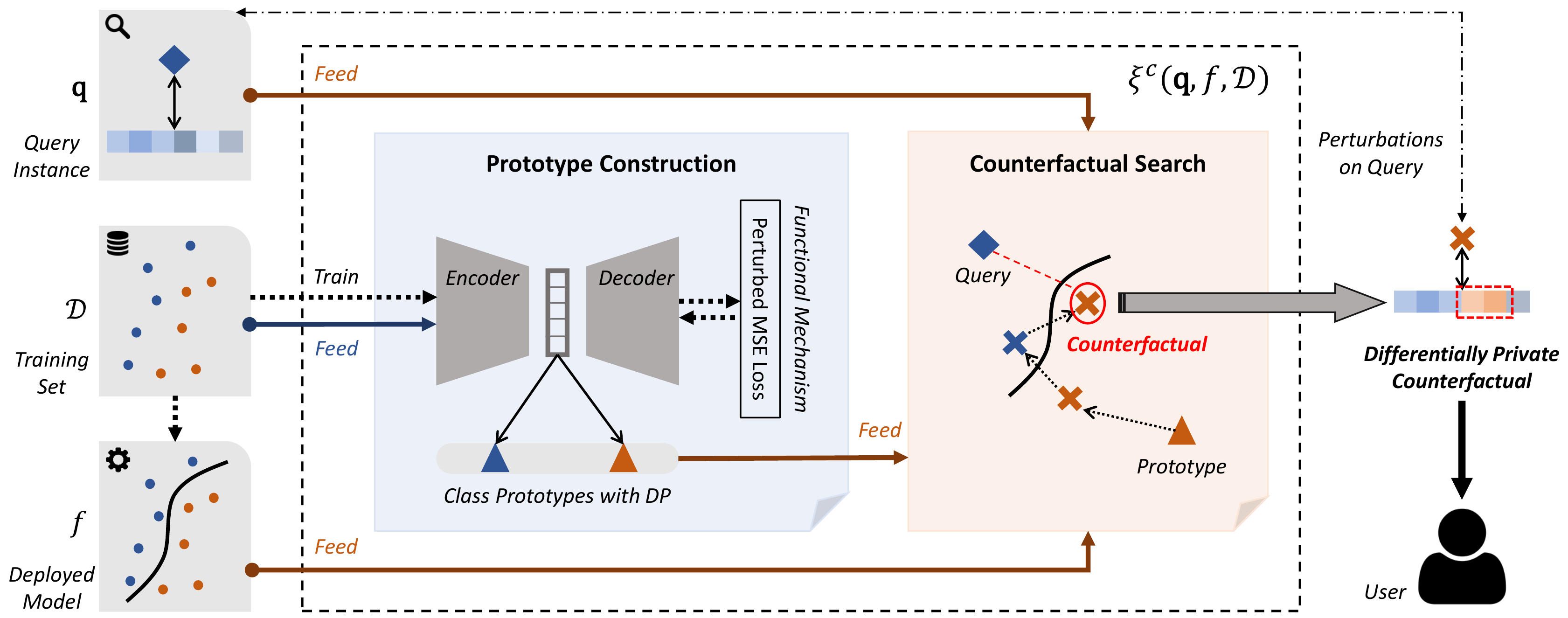} 
\caption{The designed explanation pipeline for deriving DPC.} 
\label{fig:dpc_framework} 
\end{figure*}

\subsection{Counterfactuals with Differential Privacy} 

To effectively combine the DP with counterfactuals, we first formulate the explanation process in our focused scenario. Consider a black-box model $f: \mathbb{R}^{d} \rightarrow \mathcal{R}$ trained on dataset $\mathcal{D}=\{t_1,\cdots,t_N\}$, where $t_i$ represents the $i$-th training tuple $t_i=(\mathbf{x}_{i},y_{i})$ and predicted labels are returned as model outputs. 
The counterfactual explanation can be generally indicated as a process $\xi(\mathbf{q} \ ,f)$, reflected by Eq.~\ref{eq_cf}. However, such formulation of the counterfactual process may not be sufficient to integrate the DP, since there is no way to reflect the information leakage on $\mathcal{D}$. To this end, we extend the conventional process of counterfactual explanation to $\xi^{c}(\mathbf{q} \ ,f,\mathcal{D})$ in our scenario, with an additional input dimension on $\mathcal{D}$, aiming to consider the influence of training set for counterfactual derivation. Essentially, the output of $\xi^{c}$ is a data vector in $\mathbb{R}^{d}$, following the distribution formed by training set $\mathcal{D}$. Given the deployed model $f$ as well as the domain dataset $\mathcal{D}$, counterfactual explanation is derived through the process $\xi^{c}$ with specified query $\mathbf{q}$.

We consider the scenario that adversaries could maliciously query $\xi^{c}$ with numerous points of interest, collecting obtained counterfactuals over $f$ and $\mathcal{D}$. By utilizing the collected samples, adversaries are capable of reconstructing some sensitive information about $f$ and $\mathcal{D}$. To make effective protection, we propose to deliver counterfactuals with DP guarantees (i.e., DPC), where certain noises are manually injected into $\xi^{c}$. According to the definition of DP in~\cite{dwork2014algorithmic}, we rigorously define the proposed DPC as below. 

\begin{definition}\label{def:dpc}
$\mathrm{(Differentially \ Private \ Counterfactual)}$
A counterfactual, over model $f$ and dataset $\mathcal{D}$, is said to be \textbf{$\bm{\epsilon}$-differentially private}, if the following inequality holds for any sequence of queries $\{\mathbf{q}_{1}, \cdots, \mathbf{q}_{k}\}$, any two neighbor sets $\mathcal{D}_{1},\mathcal{D}_{2}$, as well as any explanation output $\mathcal{O}_{i} \subseteq \mathbb{R}^{d}$:
\begin{equation}
    \frac{\mathrm{Pr}[\xi_{1} \in \mathcal{O}_{1}, \cdots, \xi_{k} \in \mathcal{O}_{k} ]}{\mathrm{Pr}[\xi^{\prime}_{1} \in \mathcal{O}_{1}, \cdots, \xi^{\prime}_{k} \in \mathcal{O}_{k}]} \leq \mathrm{e}^{\epsilon},
\end{equation}
where $\xi_{i}=\xi^{c}(\mathbf{q}_{i},f,\mathcal{D}_{1})$ and $\xi^{\prime}_{i}=\xi^{c}(\mathbf{q}_{i},f,\mathcal{D}_{2})$ represents the derived counterfactuals on $\mathcal{D}_{1}$ and $\mathcal{D}_{2}$ queried by $\mathbf{q}_{i}$ over $f$.
\end{definition} 

With such definition, we note that DPC generally ensures adversaries cannot obtain adequate information from the collected counterfactuals for reconstruction on $f$ and $\mathcal{D}$, up to the protection level parameterized by the budget $\epsilon$. By constraining the difference of explanation outputs over neighbor sets for $\xi^{c}$, DPC protects the sensitive information from leaking. To further derive the proposed DPC, we need to specify the explanation process $\xi^{c}$ with inputs on query $\mathbf{q}$, model $f$ and dataset $\mathcal{D}$.

\subsection{Overall Explanation Pipeline} 

We design the explanation pipeline of $\xi^{c}$ for DPC derivation, which is illustrated by Fig.~\ref{fig:dpc_framework}. Within the process, there are two major steps for calculating counterfactuals. We first construct class prototypes of $\mathcal{D}$ in the latent space with the aid of a well-trained autoencoder, where DP is effectively guaranteed by the functional mechanism through a perturbed training loss. Then, we search counterfactual samples in the latent space based on the obtained prototypes, with proper regularizations from both $\mathbf{q}$ and $f$. Specifically, the explanation process $\xi^{c}$ can be indicated as
\begin{equation}
    \xi^{c}(\mathbf{q} \ ,f,\mathcal{D}) = \xi^{c}\left(\mathbf{q} \ ,f,\xi^{p}(\mathcal{D},\psi^{\mathrm{AE}}_{\mathcal{D}})\right), 
\label{cf_process}
\end{equation}
where $\psi^{\mathrm{AE}}_{\mathcal{D}}$ denotes an autoencoder trained on $\mathcal{D}$, and $\xi^{p}$ represents the process of prototype construction. The overall DP guarantee of $\xi^{c}$ is ensured by the following Theorem~\ref{theo1}.

\begin{theorem}[DPC Immunity]\label{theo1}
    Assuming $\xi^{p}$ is $\epsilon$-differentially private in constructing class prototypes, then the counterfactual explanation process $\xi^{c}$ can also be claimed with DP under budget $\epsilon$. 
\end{theorem}
\noindent \emph{Proof Sketch --} 
For simplicity without loss of generality, we prove this theorem under a deterministic setting, but the results can be extended to randomized processes in general~\cite{pycia2015decomposing}. 
Within Eq.~\ref{cf_process}, we denote the input and output space of $\xi^{p}$ as $\mathcal{I}^{p}$ and $\mathcal{O}^{p}$. Then, the explanation process $\xi^{c}$ can be indicated as $\xi^{c}:\mathcal{I}^{c}\times \mathcal{O}^{p} \rightarrow \mathcal{O}^{c}$, where $\mathcal{I}^{c}$ represents the joint space of $\mathbf{q}$, $f$, and $\mathcal{O}^{c}$ indicates the output space of $\xi^{c}$. We fix any two neighbor sets $\mathcal{D}_{1}$, $\mathcal{D}_{2}$ differing by one sample, and any event $\mathcal{V} \subseteq \mathcal{O}^{c}$ for analysis. With notation $\mathcal{T}=\{r \in \mathcal{O}^{p}: \xi^{c}(\mathbf{q} \ ,f, r) \in \mathcal{V}\}$, we can obtain:
\begin{equation}\label{dp_proof}
    \begin{split}
        & \mathrm{Pr}\left[\xi^{c}\left(\mathbf{q} \ ,f,\xi^{p}(\mathcal{D}_{1},\psi^{\mathrm{AE}}_{\mathcal{D}_{1}})\right) \in \mathcal{V}\right] 
        = \ \ \mathrm{Pr}\left[\xi^{p}(\mathcal{D}_{1},\psi^{\mathrm{AE}}_{\mathcal{D}_{1}}) \in \mathcal{T} \right]  \\
        \leq & \ \mathrm{e}^{\epsilon} \mathrm{Pr}\left[\xi^{p}(\mathcal{D}_{2},\psi^{\mathrm{AE}}_{\mathcal{D}_{2}}) \in \mathcal{T} \right]
        = \mathrm{e}^{\epsilon} \mathrm{Pr}\left[\xi^{c}\left(\mathbf{q} \ ,f,\xi^{p}(\mathcal{D}_{2},\psi^{\mathrm{AE}}_{\mathcal{D}_{2}})\right) \in \mathcal{V}\right]. 
    \end{split}
\nonumber
\end{equation}
Thus, we demonstrate that our explanation process $\xi^{c}$ satisfies DP under budget $\epsilon$, when $\xi^{p}$ is $\epsilon$-differentially private. 

Essentially, Theorem~\ref{theo1} is a natural extension based on the post-processing immunity of DP~\cite{dwork2014algorithmic}. By injecting DP into class prototypes within $\xi^{p}$, the counterfactuals from $\xi^{c}$ are automatically ensured with DP for derivation. Thus, in the designed explanation pipeline, DPC only requires to add certain noises for $\xi^{p}$ process, and has no further requirements on the subsequent computations. In the following section, we will cover the details of prototype construction in Sec.~\ref{sec_proto} and counterfactual search in Sec.~\ref{sec_search}.

\section{Methodology} 

In this section, we first introduce our method for constructing class prototypes with DP guarantees. Then, a counterfactual searching algorithm is further introduced for DPC derivation.

\subsection{Prototype Construction}\label{sec_proto}

We construct class prototypes within the latent coding space of $\psi^{\mathrm{AE}}_{\mathcal{D}}$, which has been proved to be robust in modeling sample distances~\cite{joshi2019towards,pawelczyk2020learning,yang2021generative}. To achieve DP guarantees for counterfactuals, we need to ensure DP for the obtained prototypes from $\xi^{p}$ according to Theorem~\ref{theo1}. To this end, we specifically employ the functional mechanism to train $\psi^{\mathrm{AE}}_{\mathcal{D}}$, where a perturbed loss function is used to inject noises for the mapping process. Given a well-trained $\psi^{\mathrm{AE}}_{\mathcal{D}}$, the noisy prototypes can then be calculated by averaging the latent representations of samples in $\mathcal{D}$ over certain classes. 

Consider $\psi^{\mathrm{AE}}_{\mathcal{D}}$ as a single hidden layer network with \emph{sigmoid} activation for discussion, whose objective is to minimize the mean squared error (MSE) of reconstruction. Given an input $\mathbf{x}$, we have the following mappings for encoding and decoding in $\psi^{\mathrm{AE}}_{\mathcal{D}}$:
\begin{equation}
    \mathbf{h} = \sigma(\mathbf{W}^{\mathrm{T}}\mathbf{x}), \quad \quad \hat{\mathbf{x}} = \sigma(\mathbf{h}\mathbf{W}),
\end{equation}
where $\mathbf{h}$ is the hidden representation in latent space, $\hat{\mathbf{x}}$ denotes the reconstructed input, $\sigma(\cdot)$ indicates the \emph{sigmoid} function, and $\mathbf{W}=[\mathbf{w}_{1},\cdots,\mathbf{w}_{K}]$ represents the parameter matrix of $\psi^{\mathrm{AE}}_{\mathcal{D}}$. Then, our MSE training objective can be indicated by
\begin{equation}\label{eq_loss}
    L_{\mathcal{D}}(\mathbf{W}) = \sum
    \nolimits_{t_{i}\in \mathcal{D}} \left[\mathbf{x}_{i}-\hat{\mathbf{x}}_{i}\right]^{2} 
    = \sum
    \nolimits_{t_{i}\in \mathcal{D}} \left[\mathbf{x}_{i}-\sigma(\mathbf{h}_{i}\mathbf{W})\right]^{2}.
\end{equation}
To apply the functional mechanism in our case, we define a set of polynomial bases as below: 
\begin{equation}\label{eq_base}
    \Phi_j=\left\{g^{c_1}(\mathbf{x}, \mathbf{w}_{1}) \cdots g^{c_K}(\mathbf{x}, \mathbf{w}_{K})\Big|\sum_{m=1}^{K}c_{m}=j\right\}, \forall j \in \mathbb{N},
\end{equation}
where $g(\mathbf{x}, \mathbf{w}) = \slfrac{1}{1+\mathrm{e}^{-\sigma(\mathbf{w}^{\mathrm{T}}\mathbf{x})\mathbf{w}}}$. Since $L_{\mathcal{D}}(\mathbf{W})$ is typically continuous and differentiable, we can further express it as
\begin{equation}\label{loss_poly}
    L_{\mathcal{D}}(\mathbf{W}) = \sum_{t_{i}\in \mathcal{D}} L(t_{i},\mathbf{W}) = \sum_{t_{i}\in \mathcal{D}}\sum_{j=0}^{J}\sum_{\phi \in \Phi_j}\lambda_{\mathbf{x}_i}^{\phi}\phi(\mathbf{x}_i, \mathbf{W}), 
\end{equation}
based on the Stone-Weierstrass theorem~\cite{rudin1964principles}, where $\lambda_{\mathbf{x}_i}^{\phi}$ indicates the coefficient of polynomial base $\phi$. Given Eq.~\ref{loss_poly}, functional mechanism perturbs $L_{\mathcal{D}}(\mathbf{W})$ by adding Laplace noises into $\lambda_{\mathbf{x}_i}^{\phi}$, and further derives $\widetilde{\mathbf{W}}$ minimizing the perturbed loss $\widetilde{L}_{\mathcal{D}}(\mathbf{W})$. The DP of such process is guaranteed by the following lemma and theorem.

\begin{lemma}[Sensitivity Upper Bound]\label{lemma1}
Assume $\mathcal{D}_1$ and $\mathcal{D}_2$ are two arbitrary neighbor sets. Let $L_{\mathcal{D}_1}(\mathbf{W})$ and $L_{\mathcal{D}_2}(\mathbf{W})$ respectively be the training objectives on $\mathcal{D}_1$ and $\mathcal{D}_2$. Then, we have the following upper bound for global sensitivity:
\begin{equation}
    \sum_{j=0}^{J}\sum_{\phi\in\Phi_j} \Bigg\lVert\sum_{t_i\in \mathcal{D}_1}\lambda_{\mathbf{x}_i}^{\phi} - \sum_{t_i^{\prime}\in \mathcal{D}_2}\lambda_{\mathbf{x}_i^{\prime}}^{\phi}\Bigg\rVert_{1} 
    \leq 2\max_{\mathbf{x}} \sum_{j=0}^{J}\sum_{\phi\in\Phi_j}\Big\lVert\lambda_{\mathbf{x}}^{\phi}\Big\rVert_{1} \ , 
\end{equation}
where $\lambda_{\mathbf{x}}^{\phi}$ indicates the polynomial coefficients of training objectives. 
\end{lemma}
\noindent \emph{Proof Sketch --} 
We assume $\mathcal{D}_1$ and $\mathcal{D}_2$ are the neighbor sets differing by the last tuple, without loss of generality. Let $t_{N}$ and $t_{N}^{\prime}$ respectively be the last tuple in $\mathcal{D}_1$ and $\mathcal{D}_2$. Then, we have 
\begin{equation}
\begin{split}
    & \sum_{j=0}^{J}\sum_{\phi\in\Phi_j} \Bigg\lVert\sum_{t_i\in \mathcal{D}_1}\lambda_{\mathbf{x}_i}^{\phi} - \sum_{t_i^{\prime}\in \mathcal{D}_2}\lambda_{\mathbf{x}_i^{\prime}}^{\phi}\Bigg\rVert_{1}  
    = \  \sum_{j=0}^{J}\sum_{\phi\in\Phi_j}\Big\lVert\lambda_{\mathbf{x}_{N}}^{\phi}-\lambda_{\mathbf{x}_{N}^{\prime}}^{\phi}\Big\rVert_{1}  \\
    \leq \ & \sum_{j=0}^{J}\sum_{\phi\in\Phi_j} \Big\lVert\lambda_{\mathbf{x}_{N}}^{\phi}\Big\rVert_{1} + \sum_{j=0}^{J}\sum_{\phi\in\Phi_j} \Big\lVert\lambda_{\mathbf{x}_{N}^{\prime}}^{\phi}\Big\rVert_{1} 
    \leq \ 2\max_{\mathbf{x}} \sum_{j=0}^{J}\sum_{\phi\in\Phi_j}\Big\lVert\lambda_{\mathbf{x}}^{\phi}\Big\rVert_{1}. 
\end{split}
\nonumber 
\end{equation}

\begin{theorem}[DP Guarantee]\label{theo2}
Let the objective sensitivity be $\Delta=2\max_{\mathbf{x}} \sum_{j=0}^{J}\sum_{\phi\in\Phi_j}\lVert\lambda_{\mathbf{x}}^{\phi}\rVert_{1}$. Then, the $\epsilon$-DP of the process can be ensured by a perturbed objective $\widetilde{L}_{\mathcal{D}}(\mathbf{W})=\sum_{j=0}^{J}\sum_{\phi\in\Phi_j}\widetilde{\lambda}^{\phi}\phi(\mathbf{W})$, in which perturbed coefficients are calculated as
\begin{equation}\label{pert_coef}
    \widetilde{\lambda}^{\phi} = \sum_{t_{i}\in \mathcal{D}} \lambda_{\mathbf{x}_i}^{\phi} + \mathrm{Lap}\left(\frac{\Delta}{\epsilon}\right),
\end{equation}
with scaled Laplace noises parameterized by $\Delta/\epsilon$. 
\end{theorem}
\noindent \emph{Proof Sketch --} 
We similarly assume that $\mathcal{D}_1$ and $\mathcal{D}_2$ differ by the last tuple, i.e., $t_{N}$ and $t_{N}^{\prime}$, respectively. Given $\Delta$ and $\widetilde{L}_{\mathcal{D}}(\mathbf{W})$ as stated, we can make the following derivations:
\begin{equation}
\begin{split}
    & \frac{\mathrm{Pr}\left(\widetilde{L}_{\mathcal{D}_{1}}(\mathbf{W})\right)}{\mathrm{Pr}\left(\widetilde{L}_{\mathcal{D}_{2}}(\mathbf{W})\right)}
    = \ \frac{\prod_{j=0}^{J}\prod_{\phi\in\Phi_j}\mathrm{e}^{\frac{\epsilon \big\lVert\sum_{t_i\in \mathcal{D}_{1}}\lambda_{\mathbf{x}_i}^{\phi}-\widetilde{\lambda}^{\phi}\big\rVert_{1}}{\Delta}}}{\prod_{j=0}^{J}\prod_{\phi\in\Phi_j}\mathrm{e}^{\frac{\epsilon \Big\lVert\sum_{t_{i}^{\prime}\in \mathcal{D}_{2}}\lambda_{\mathbf{x}_{i}^{\prime}}^{\phi}-\widetilde{\lambda}^{\phi}\Big\rVert_{1}}{\Delta}}}  \\
    \leq \ & \prod_{j=0}^{J}\prod_{\phi\in\Phi_j} \mathrm{e}^{\frac{\epsilon}{\Delta}\Big\lVert\sum_{t_i\in \mathcal{D}_{1}}\lambda_{\mathbf{x}_{i}}^{\phi}-\sum_{t_{i}^{\prime}\in \mathcal{D}_{2}}\lambda_{\mathbf{x}_{i}^{\prime}}^{\phi}\Big\rVert_{1}} 
    = \ \prod_{j=0}^{J}\prod_{\phi\in\Phi_j}\mathrm{e}^{\frac{\epsilon}{\Delta}\Big\lVert\lambda_{\mathbf{x}_{N}}^{\phi}-\lambda_{\mathbf{x}_{N}^{\prime}}^{\phi}\Big\rVert_{1}} \\
    = \ & \mathrm{e}^{\frac{\epsilon}{\Delta}\sum_{j=0}^{J}\sum_{\phi\in\Phi_j}\Big\lVert\lambda_{\mathbf{x}_{N}}^{\phi}-\lambda_{\mathbf{x}_{N}^{\prime}}^{\phi}\Big\rVert_{1}}
    \leq \ \mathrm{e}^{\frac{\epsilon}{\Delta} \cdot 2\max_{\mathbf{x}} \sum_{j=0}^{J}\sum_{\phi\in\Phi_j}\lVert\lambda_{\mathbf{x}}^{\phi}\rVert_{1}}
    = \ \mathrm{e}^{\epsilon}.
\end{split}
\nonumber 
\end{equation}
Thus, we obtain that $\slfrac{\mathrm{Pr}\left(\widetilde{L}_{\mathcal{D}_{1}}(\mathbf{W})\right)}{\mathrm{Pr}\left(\widetilde{L}_{\mathcal{D}_{2}}(\mathbf{W})\right)} \leq \mathrm{e}^{\epsilon}$, which proves that $\epsilon$-DP is achieved with the perturbed objective $\widetilde{L}_{\mathcal{D}}(\mathbf{W})$.

With Eq.~\ref{eq_loss} and Eq.~\ref{eq_base}, we obtain the following objective for $\psi^{\mathrm{AE}}_{\mathcal{D}}$:
\begin{equation}\label{case_loss}
\begin{split}
    L_{\mathcal{D}}(\mathbf{W}) 
    = & \sum\nolimits_{t_{i}\in \mathcal{D}}\left(\mathbf{x}_{i}^{2}\right)-\sum\nolimits_{p=1}^{K}\left(2\sum\nolimits_{t_{i}\in \mathcal{D}}\mathbf{x}_{i} \cdot g\left(\mathbf{x}_{i}, \mathbf{w}_{p}\right) \right)  \\
    & + \sum\nolimits_{p \geq 1, \ q \leq K} \left(\sum\nolimits_{t_{i}\in \mathcal{D}} g\left(\mathbf{x}_{i}, \mathbf{w}_{p}\right) \cdot g\left(\mathbf{x}_{i}, \mathbf{w}_{q}\right) \right). 
\end{split}
\end{equation}
Assuming input values are normalized into $[-1,1]$, we can further relax the sensitivity upper bound in Lemma~\ref{lemma1} based on Eq.~\ref{case_loss} as
\begin{equation}\label{case_bound}
    \Delta \leq
    2\max_{\mathbf{x}}\left(\mathbf{x}^{2}+\sum_{p=1}^{K}2\mathbf{x}_{(p)}+1\right) \leq 4(K+1),
\end{equation}
where $\mathbf{x}_{(p)}$ denotes the $p$-th dimension of an arbitrary input vector. Following Theorem~\ref{theo2} with Eq.~\ref{case_bound}, $\psi^{\mathrm{AE}}_{\mathcal{D}}$ is built with DP guarantees. The class prototypes in the latent space can then be derived by 
\begin{equation}\label{prot_construct}
    \rho_{s} = \frac{1}{N_s}\sum\nolimits_{i=1}^{N_s}\psi^{\mathrm{Enc}}(\mathbf{x}_{i}), \quad \mathbf{x}_{i} \in \{(\mathbf{x}_{i},y_{i}):y_{i}=s\},
\end{equation}
where $\rho_{s}$ denotes the prototype for class $s$, $N_s$ is the total number of instances in class $s$, and $\psi^{\mathrm{Enc}}$ represents the encoder part of $\psi^{\mathrm{AE}}_{\mathcal{D}}$.

\subsection{Counterfactual Search}\label{sec_search}

With the DP-guaranteed prototypes derived, we further search counterfactuals for model explanations. Basically, for class $s$, we aim to find hypothetical sample $\psi^{\mathrm{Dec}}(\rho_{s}+\delta)\in \mathbb{R}^{d}$ that satisfies counterfactual properties~\cite{wachter2017counterfactual}, where $\psi^{\mathrm{Dec}}$ indicates the decoder part of $\psi^{\mathrm{AE}}_{\mathcal{D}}$. We consider the following properties for searching. 

\smallskip

\textbf{\emph{Prediction}}. We prefer the counterfactuals for explanation which have predicted labels as desired. Thus, we design a loss term based on the cross-entropy as below to search counterfactuals in class $s$
\begin{equation}
    L_{\mathrm{pred}} = -\sum\nolimits_{n}\mathbf{y}_{(n)}\log\left[f\left(\psi^{\mathrm{Dec}}(\rho_{s}+\delta)\right)\right]_{(n)}, \quad \mathbf{y}_{(s)}=1, 
\end{equation}
where $\mathbf{y}$ is a one-hot vector indicating the desired output class. Essentially, $L_{\mathrm{pred}}$ encourages the target prediction for the perturbed samples, so that the searched counterfactual can reflect the changes across decision boundaries. 

\smallskip

\textbf{\emph{Distance}}. We prefer the counterfactuals that are close to the query $\mathbf{q}$. In particular, we use the $L_{2}$ norm to measure the sample distance in data space, and design a loss term accordingly as
\begin{equation}
    L_{\mathrm{dist}} = \Big\lVert \ \psi^{\mathrm{Dec}}(\rho_{s}+\delta) - \mathbf{q} \ \Big\rVert_{2}. 
\end{equation}
With the $L_{\mathrm{dist}}$ term, the searching process would attach more preferences to those samples that are similar to $\mathbf{q}$, ensuring the minimal changes to flip the model outcomes.  

\smallskip

\textbf{\emph{Prototype}}. We also prefer the encoding of perturbed samples in latent space to be close to the class prototypes, which would guide the searched counterfactual towards an in-distribution sample. Specifically, we employ the $L_{2}$ norm as well to measure distances in latent space, and the related loss can be indicated by
\begin{equation}
    L_{\mathrm{prot}} = \big\lVert \ (\rho_{s}+\delta) - \rho_{s} \ \big\rVert_{2} = \big\lVert \ \delta \ \big\rVert_{2}.
\end{equation}
The $L_{\mathrm{prot}}$ term explicitly regularizes the perturbation $\delta$ towards the target class prototype, and speeds up the search with the average encoding of target class $s$. 

Considering these aspects, we can then obtain the overall objective function for counterfactual search as follows.
\begin{equation}\label{search_obj}
    L_{cs}(\delta) = \alpha L_{\mathrm{pred}} + \beta L_{\mathrm{dist}} + \gamma L_{\mathrm{prot}},
\end{equation}
where $\alpha$, $\beta$ and $\gamma$ are the balancing coefficients. By minimizing $L_{cs}$, proper perturbations can be derived, and DPC is further obtained through the decoder $\psi^{\mathrm{Dec}}$. Within such searching process, it is also proved that our derived DPC will not introduce extra bias from the noisy prototypes. The robustness of DPC search is guaranteed by the following Theorem~\ref{theo3}. 

\begin{theorem}[Unbiased DPC Searching]\label{theo3}
    The searching process of DPC does not introduce extra bias, i.e., 
    \begin{equation}
        \mathrm{Bias}[\check{cs}(\rho)] \coloneqq \mathbb{E}_{\rho}[\mathrm{Diff}(\check{cs}(\rho))] = \mathbf{0},
    \end{equation}
    where $\mathrm{Diff}(\cdot)$ denotes the difference from the true value, $\check{cs}(\cdot)$ indicates the counterfactual search process, $\rho$ represents the noisy class prototypes, and the expectation is taken over the noisy distribution. 
\end{theorem}
\noindent \emph{Proof Sketch --}
Let $\mathrm{Pd}_{\rho}(\cdot)$ denote the probability density function of $\rho$. According to Eq.~\ref{pert_coef}, we know $\mathrm{Pd}_{\rho}$ is symmetric regarding to the prototypes without DP noises. Then, the expectation of the resulting difference can be calculated as below. 
\begin{equation}
    \begin{split}
    & \mathbb{E}_{\rho}[\mathrm{Diff}(\check{cs}(\rho))] =  \int_{\mathbf{z}}\mathrm{Diff}(\check{cs}(\mathbf{z})) \cdot \mathrm{Pd}_{\rho}(\mathbf{z}) \ \mathrm{d}\mathbf{z} \\
    = & \ \frac{1}{2} \int_{\mathbf{z}} \left[\mathrm{Diff}(\check{cs}(\mathbf{z}))+\mathrm{Diff}(\check{cs}(\mathrm{Ref}_{\rho}(\mathbf{z})))\right] \mathrm{Pd}_{\rho}(\mathbf{z}) \ \mathrm{d}\mathbf{z} \\
    = & \ \frac{1}{2} \int_{\mathbf{z}} \mathbf{0} \cdot \mathrm{Pd}_{\rho}(\mathbf{z}) \ \mathrm{d}\mathbf{z}
    = \ \mathbf{0}, 
    \end{split}
\nonumber 
\end{equation}
where $\mathrm{Ref}_{\rho}(\cdot)$ indicates a reflection operator~\cite{coxeter1961introduction} across $\rho$ that is commutative with the counterfactual search process.

\begin{algorithm}[t]

\KwIn{Dataset $\mathcal{D}$, Privacy Budget $\epsilon$}
\KwOut{Noisy Class Prototype $\rho_{s}$, $\forall s$}

\medskip

- Set $\Delta$ for scaled Laplace noises according to Eq.~\ref{case_bound};  \\

- \For{each $0 \leq j \leq J$}{
    \For{each $\phi \in \Phi_{j}$}{
      Perturb coefficients based on Eq.~\ref{pert_coef};
    }
}

- Construct the perturbed objective $\widetilde{L}_{\mathcal{D}}(\mathbf{W})$ by Theo.~\ref{theo2};  \\

- Train autoencoder $\psi^{\mathrm{AE}}_{\mathcal{D}}$ with $\widetilde{L}_{\mathcal{D}}(\mathbf{W})$;  \\

- Derive the noisy prototype for class $s$ based on Eq.~\ref{prot_construct}. 

\caption{Prototype Construction with DP Guarantees}
\label{algo1}
\end{algorithm}

\begin{algorithm}[t]

\KwIn{Prototype $\rho_{s}$, Query $\mathbf{q}$, Autoencoder $\psi^{\mathrm{AE}}_{\mathcal{D}}$, Model $f$}
\KwOut{Counterfactual Sample $\xi$}

\medskip

- Select the target class $s \ (\forall s)$ for counterfactual reasoning;  \\

- Set the hyper-parameter $\alpha$, $\beta$, $\gamma$ for Eq.~\ref{search_obj};  \\

- \For{each iteration}{
    Optimize the perturbation $\delta$ in objective Eq.~\ref{search_obj};  \\
}

- Return the counterfactual obtained by $\psi^{\mathrm{Dec}}(\rho_{s}+\delta)$. 

\caption{Counterfactual Search for Explanation}
\label{algo2}
\end{algorithm}

\subsection{Implemented Algorithms} 

We summarize our methods in Algorithm~\ref{algo1} and Algorithm~\ref{algo2}. In practice, to obtain a robust latent space, the employed autoencoder $\psi^{\mathrm{AE}}_{\mathcal{D}}$ may have more than one hidden layer for encoding, which could make our sensitivity upper bound derived in Eq.~\ref{case_bound} fail on DP guarantees. To handle this, we specifically add \emph{normalization layer} before each hidden layer, so as to ensure the input values for encoding fall into the assumed range $[-1,1]$. Throughout the process, it is noted that the major time consumption comes from the searching in Algorithm~\ref{algo2}, since Algorithm~\ref{algo1} can be executed offline when dataset $\mathcal{D}$ is given. The overall time complexity for explanation mainly depends on the \emph{iteration steps} for minimizing Eq.~\ref{search_obj}, which is the same as the most methods for counterfactuals~\cite{mothilal2020explaining,moore2019explaining,white2019measurable}. Thus, the proposed DPC does not involve extra computation overhead during explanation, which makes it easier to be implemented into the practical pipelines.

\section{Experiments}     

In this section, we comprehensively evaluate the proposed DPC framework in real-world datasets from different perspectives. Overall, we aim to answer the following key research questions. 

\begin{itemize}[leftmargin=*]

    \item Can we effectively inject the controlled noises for DP guarantees of counterfactuals through the proposed framework? 
    
    \item How effective is the derived DPC for model explanation, compared with the existing counterfactual methods? 
    
    \item Is the derived DPC able to relieve the risks on model security and data privacy when explanations are maliciously collected? 

\end{itemize}

\subsection{Evaluation on Noise Injection for DP} 

In this part, we evaluate the noise injection of the proposed DPC framework. Essentially, our evaluation focuses on the autoencoder $\psi^{\mathrm{AE}}_{\mathcal{D}}$ trained with the functional mechanism, which plays a key role for the DPC derivation process. 

\subsubsection{Experimental Settings.} 

\begin{table}[t] 
\small 
\caption{Data statistics in experiments.}
\vspace{-0.4cm}
\centering
\setlength{\tabcolsep}{3.9pt} 
\begin{normalsize}
\begin{tabular}[width=1cm]{cccccc}
\toprule
Dataset        & \#Instance      & \#Feature    & \# Class  & Type  \\
\midrule
Adult       & $48,842$     & $24$  &  $2$   & Mixed   \\
Hospital      & $101,766$     & $127$  &  $2$   & Mixed   \\
HomeCredit      & $344,971$     & $39$  &  $2$   & Mixed  \\
MNIST    & $70,000$     & $28\times28$  &  $10$   & Image   \\
Purchase    & $197,324$     & $600$  &  $100$   & Binary   \\
Texas    & $65,761$     & $3,622$  &  $100$   & Binary   \\
\bottomrule
\end{tabular}
\end{normalsize}
\label{tab:data}
\end{table}

We consider six real-world datasets with different types for evaluation, and the data statistics are specifically shown in Table~\ref{tab:data}. 

\begin{itemize}[leftmargin=*]
\item \textbf{\textit{Adult}}: This is a real-world dataset for the annual income prediction~\cite{dua2017uci}, where each instance is labelled as ``>50K'' or ``<=50K''. The task is to predict the income level for a given adult profile. 

\item \textbf{\textit{Hospital}}: This dataset contains information on diabetic patients from $130$ hospitals in the US~\cite{strack2014impact}. The overall prediction task is to forecast the patient readmission status within $30$ days. 

\item \textbf{\textit{HomeCredit}}: This is a real-world business dataset for the client risk assessment, where the goal is to predict clients' repayment abilities for given loans~\cite{al2019comparison,yang2021model}. 

\item \textbf{\textit{MNIST}}: This is a simple image set of hand-written digits~\cite{yannmnist}, which is used as a benchmark for multi-class scenarios. The data includes $10$ different labels, ranging from $0$ to $9$. 

\item \textbf{\textit{Purchase}}: This is another real-world business dataset, whose goal is to predict the customer responses to offers and discounts based on the shopping history\cite{shokri2017membership}.

\item \textbf{\textit{Texas}}: This dataset is about the patient status at different health facilities~\cite{nasr2018machine}, which is collected by the Texas Department of State Health Services. The goal of this data is to predict the patients' primary procedures based on their remaining attributes. 
\end{itemize} 

\noindent In our experiments, we employ different autoencoder architectures for different datasets. Specifically, for the mixed datasets (i.e., Adult, Hospital, HomeCredit), our autoencoders consist of $2$ dense layers in the encoder part with respectively $32$, $16$ neurons for each layer. As for the image dataset (i.e., MNIST), instances are typically flattened as pixel features, and we build the autoencoder with $4$ dense layers for encoding, where each layer has $256$, $128$, $64$, $32$ neurons, respectively. Similarly for the binary datasets (i.e., Purchase, Texas), we use $5$ dense layers for feature encoding, and layers are respectively designed with $512$, $256$, $128$, $64$, $32$ neurons. In general, all the decoders follow the same architecture of the corresponding encoder in a reversed order. 
To effectively calculate the sensitivity upper bound of $\psi^{\mathrm{AE}}_{\mathcal{D}}$ by Eq.~\ref{case_bound}, we set $K$ as the maximal number of neurons over all hidden layers. 
The autoencoders are typically trained with Adam optimizer for $300$, $30$, $500$ epochs in batch size $128$, $256$, $256$, respectively, on the mixed, image, binary datasets. 

\subsubsection{Quantitative Results.} 

\begin{figure}[t] 
\centering 
\includegraphics[width=\columnwidth]{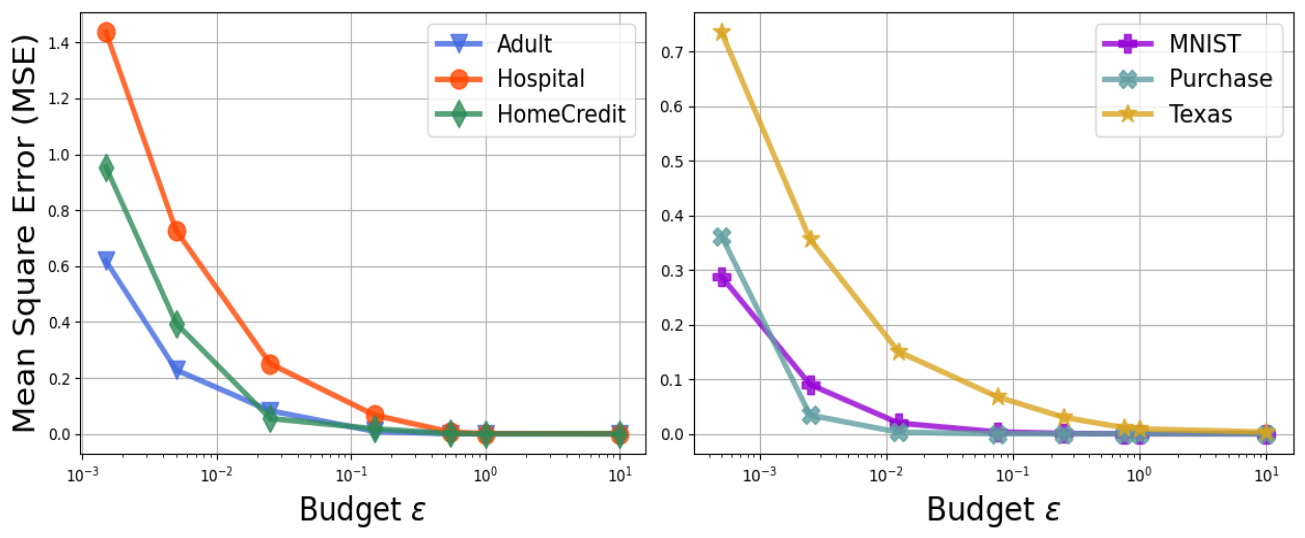} 
\vspace{-0.6cm}
\caption{The MSE of $\psi^{\mathrm{AE}}_{\mathcal{D}}$ over $\epsilon$ on different datasets.} 
\label{fig:mse_dp} 
\end{figure} 

We show the reconstruction MSE of $\psi^{\mathrm{AE}}_{\mathcal{D}}$ under different budget $\epsilon$ in Fig.~\ref{fig:mse_dp}. According to the empirical results, we observe a significant trade-off between the DP budget and model performance over all datasets, where the performance degrades (i.e., higher MSE) when the budget $\epsilon$ becomes lower. Reflected by Eq.~\ref{pert_coef}, we know that lower $\epsilon$ generally indicates a stronger noise injection, which sacrifices the reconstruction quality of $\psi^{\mathrm{AE}}_{\mathcal{D}}$ for stronger DP guarantees. Our observation here is consistent with many research work~\cite{pannekoek2021investigating,mochaourab2021robust} on DP, further demonstrating that the functional mechanism we employ for $\psi^{\mathrm{AE}}_{\mathcal{D}}$ training works effectively on noise injection. From the results in Fig.~\ref{fig:mse_dp} , we also note different data have different degradation ratio against DP, which depends on the feature dimensions and model architectures. 

\subsubsection{Qualitative Cases.} 

\begin{figure}[t] 
\centering 
\includegraphics[width=\columnwidth]{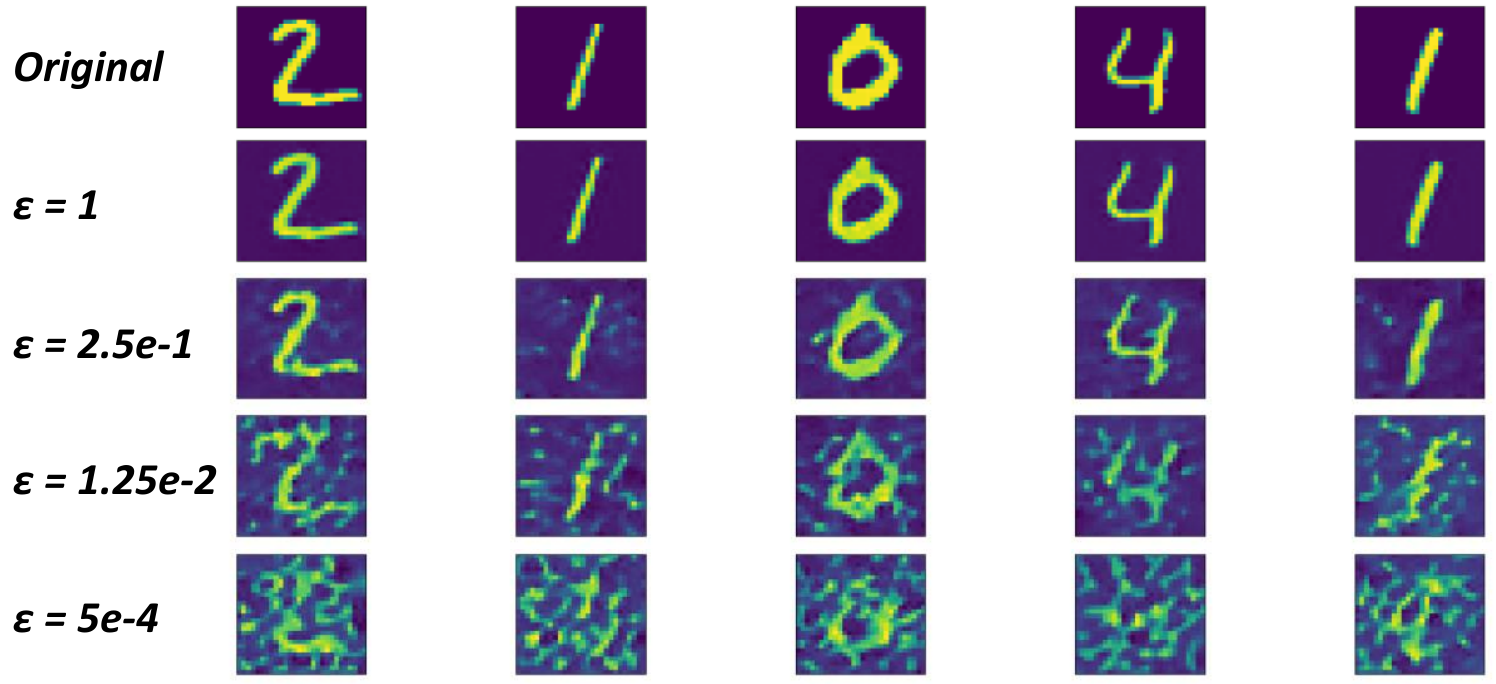} 
\vspace{-0.5cm}
\caption{The case results of $\psi^{\mathrm{AE}}_{\mathcal{D}}$ on MNIST over different $\epsilon$.} 
\label{fig:dpae_case} 
\end{figure} 

To have qualitative evaluations on noise injection, we further visualize some case results of $\psi^{\mathrm{AE}}_{\mathcal{D}}$ on MNIST dataset over different budget $\epsilon$. As shown in Fig~\ref{fig:dpae_case}, we note that DP noises, parameterized by $\epsilon$, are effectively injected with our proposed framework. When $\epsilon$ is large (e.g., $\epsilon=1$), it is observed that the original inputs can be well reconstructed with a reasonable visual quality. When $\epsilon$ becomes gradually smaller, we observe significant degradation on reconstruction performance of $\psi^{\mathrm{AE}}_{\mathcal{D}}$. In the extreme case where $\epsilon$ is pretty small (e.g., $\epsilon$=5e-4), the reconstructed digits can hardly be identified with the strong DP noises injected. Throughout this set of case studies, we can clearly find the trade-off between DP and performance, where stronger DP typically leads to a weaker performance with more noises injected.

\subsection{Evaluation on Derived Counterfactuals} 

In this part, we evaluate the counterfactuals obtained from the proposed DPC framework, and compare its effectiveness as well as efficiency for explanation with the existing methods. 

\subsubsection{Experimental Settings.} 

We select four existing counterfactual explanation methods as baselines for comparison, and all involved methods are set up with their default hyperparameter settings. 

\begin{itemize}[leftmargin=*]
\item \textbf{\textit{DiCE}}~\cite{mothilal2020explaining}: This method generates diverse counterfactuals by providing feature-perturbed samples from the query, where the specific perturbations are computed through an iterative manner. 

\item \textbf{\textit{CADEX}}~\cite{moore2019explaining}: This method employs the gradient-based algorithm to iteratively perturb the query for counterfactual explanation, which is a direct application of adversarial attack methods. 

\item \textbf{\textit{C-CHVAE}}~\cite{pawelczyk2020learning}: This method pre-train an autoencoder to transform the original space into a latent space, and then perturbs the latent representation of query for counterfactual explanation. 

\item \textbf{\textit{REVISE}}~\cite{joshi2019towards}: This method similarly uses the generative model to capture the underlying data manifold, aiming to generate the smallest set of changes for altering model outcomes. 
\end{itemize}

\noindent Besides, in our experiments, we simply consider the neural network as our target classification model $f$ for explaining. For different datasets, we employ different neural architectures for prediction. Specifically, for the mixed and binary tabular datasets, we use the fully-connected neural network with $\mathrm{tanh}$ activations. As for the image dataset, we use the convolutional neural network with ReLU activations. 
Regarding to the counterfactual searching on $f$, we have different sets of balancing coefficients for Eq.~\ref{search_obj}. For the mixed datasets, we fix the hyperparameters as $\alpha=1$, $\beta=0.5$, $\gamma=0.1$. For the image dataset, the hyperparameters are set as $\alpha=1$, $\beta=0.2$, $\gamma=20$. For the binary datasets, we set $\alpha=1$, $\beta=0.5$, $\gamma=10$.

\begin{table*}[t]
\small
\renewcommand\arraystretch{1.2} 
\centering 
\caption{The flipping ratio (FR) of derived counterfactuals over different explanation methods.} 
\vspace{-0.3cm}
\label{tab:fr} 
\begin{tabular}{|p{2cm}||p{0.9cm}p{0.9cm}|p{0.9cm}p{0.9cm}|p{0.9cm}p{0.9cm}|p{0.9cm}p{0.9cm}|p{0.9cm}p{0.9cm}|p{0.9cm}p{0.9cm}|}
\hline
\multirow{2}{*}{\textit{\begin{tabular}[c]{@{}l@{}}Method\\ Dataset\end{tabular}}} & \multicolumn{2}{c|}{\textbf{DiCE}} & \multicolumn{2}{c|}{\textbf{CADEX}} & \multicolumn{2}{c|}{\textbf{C-CHVAE}} & \multicolumn{2}{c|}{\textbf{REVISE}} & \multicolumn{2}{c|}{\textbf{DPC ($\epsilon=0.025$)}} & \multicolumn{2}{c|}{\textbf{DPC ($\epsilon=0.75$)}} \\ \cline{2-13} 
                                                                                   & \textit{n/spec.}  & \textit{spec.} & \textit{n/spec.}  & \textit{spec.}  & \textit{n/spec.}   & \textit{spec.}   & \textit{n/spec.}   & \textit{spec.}  & \textit{n/spec.}    & \textit{spec.}    & \textit{n/spec.}    & \textit{spec.}   \\ \hline
\textbf{Adult}                                                                     & 0.972             & /              & 0.926             & /               & 0.946              & /                & 0.902              & /               & 0.918               & /                 & 0.968               & /                \\
\textbf{Hospital}                                                                  & 0.892             & /              & 0.810             & /               & 0.838              & /                & 0.788              & /               & 0.826               & /                 & 0.884               & /                \\
\textbf{HomeCredit}                                                                & 0.902             & /              & 0.878             & /               & 0.862              & /                & 0.832              & /               & 0.868               & /                 & 0.908               & /                \\
\textbf{MNIST}                                                                     & 0.812             & 0.204          & 0.724             & 0.138           & 0.788              & 0.412            & 0.682              & 0.316           & 0.756               & 0.720             & 0.790               & 0.742            \\
\textbf{Purchase}                                                                  & 0.778             & 0.056          & 0.744             & 0.028           & 0.822              & 0.126            & 0.734              & 0.166           & 0.722               & 0.382             & 0.772               & 0.406            \\
\textbf{Texas}                                                                     & 0.656             & 0.018          & 0.624             & 0.012           & 0.728              & 0.104            & 0.764              & 0.138           & 0.706               & 0.370             & 0.736               & 0.402            \\ \hline
\end{tabular}
\end{table*}

\subsubsection{Counterfactual Effectiveness.}

\begin{table}[t]
\small
\centering 
\caption{The average distance (AD) of counterfactuals.} 
\vspace{-0.3cm}
\label{tab:ad} 
\begin{tabular}{l||cccc|c}
                    & \textbf{DiCE} & \textbf{CADEX} & \textbf{C-CHVAE} & \textbf{REVISE} & \textbf{DPC} \\ \hline
\textbf{Adult}      & 1.32         & 0.61              & 1.66                & 1.61               & 1.62             \\
\textbf{Hospital}   & 2.51         & 1.54              & 2.68                & 2.70               & 2.63             \\
\textbf{HomeCredit} & 2.02         & 1.23              & 2.27                & 2.19               & 2.25             \\
\textbf{MNIST}      & 2.89        & 1.81              & 3.04                & 3.09               & 2.99             \\
\textbf{Purchase}   & 2.90             & 1.70              & 3.20                & 3.30               & 3.00             \\
\textbf{Texas}      & 2.50          & 2.10              & 2.70                & 2.60               & 2.60            
\end{tabular}
\end{table}

We first employ the \emph{Flipping Ratio} (FR) as the metric to evaluate the effectiveness of derived counterfactuals. In particular, FR can be calculated as follows:
\begin{equation}
    \mathrm{FR}=\left|\mathcal{X}_{f}\right| \Big/ \left|\mathcal{X}_{q}\right|,
\end{equation}
where $\mathcal{X}_{f}$ indicates the set of derived counterfactuals which can flip the model outcomes, and $\mathcal{X}_{q}$ represents the whole set of query instances for explanation. To make a fair comparison, we only derive one counterfactual sample per query from each explanation method. In experiments, we have $500$ testing queries in total (i.e., $\left|\mathcal{X}_{q}\right|=500$), which are randomly selected from the test set. Table~\ref{tab:fr} shows our experimental results over all datasets. From the results, we note that the proposed DPC can achieve a competitive FR performance compared with the existing counterfactual methods. When DP budget is sufficient (i.e., lower DP requirement), the derived DPC can even outperform most of the baselines on FR. Besides, we also observe a significant improvement of DPC in \emph{spec.}
\footnote{\textit{spec.} denotes the counterfactual scenarios where a specified prediction label is given, while \textit{n/spec.} denotes the scenarios without specified labels. For binary classification, \emph{spec.} equals to \textit{n/spec.} since there is only one possibility for altering model outcomes.} 
scenarios, which may result from the fact that DPC searching is started from the specified class prototype instead of the query as the baselines.
Then, we employ the \emph{Average Distance} (AD) to further evaluate the counterfactual effectiveness. Specifically, AD can be computed as:
\begin{equation}
 \mathrm{AD} = \sum\nolimits_{\mathbf{\xi} \in \mathcal{S}^{cf}}\mathrm{Euclid\_Distance}(\mathbf{q}, \mathbf{\xi}) \Big/ \left|\mathcal{S}^{cf}\right|,
\end{equation}
where $\mathcal{S}^{cf}$ denotes the set of derived counterfactuals regarding $\mathbf{q}$. In experiments, we set $|\mathcal{S}^{cf}|=10$, and compare AD among different counterfactual methods over different datasets. The relevant results are shown in Table~\ref{tab:ad}, where DPC is evaluated under $\epsilon=0.025$. From the results, we note that DPC may not be as close to the query as DiCE and CADEX when injected with certain noises, but its overall AD performance is still competitive to C-CHVAE and REVISE.

\subsubsection{Counterfactual Efficiency.} 

\begin{figure}[t] 
\centering 
\includegraphics[width=0.9\columnwidth]{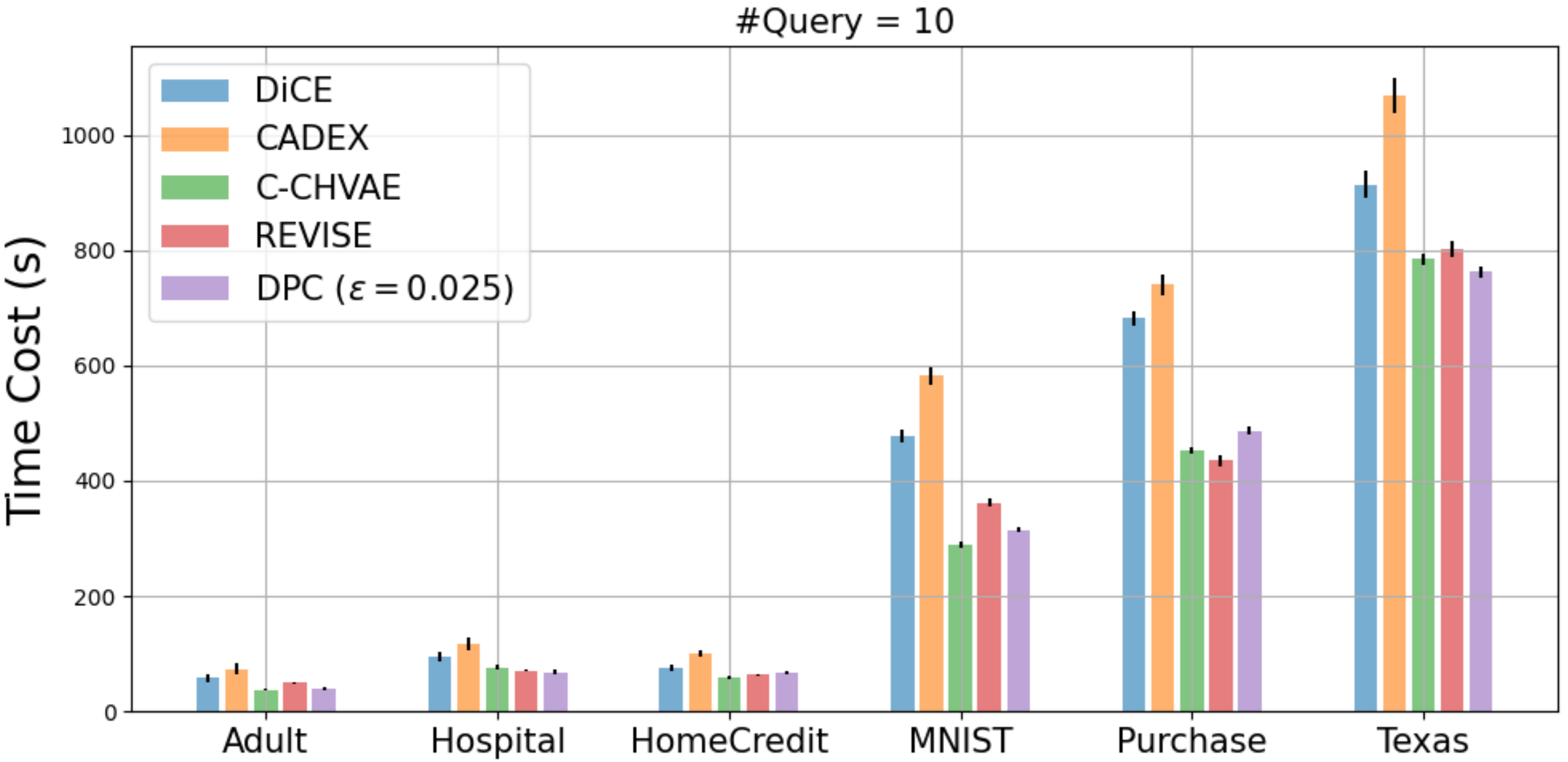} 
\vspace{-0.25cm}
\caption{The time cost comparison for different methods.} 
\label{fig:time} 
\end{figure}

We compare the time consumption of counterfactual derivation among different methods. In experiments, we only count the time cost of explanation process, and test the sample derivation for $10$ different queries in each run. The final results, shown in Fig.~\ref{fig:time}, are reported by averaging $5$ runs over different set of input queries. From Fig.~\ref{fig:time}, we observe that our proposed DPC does not introduce extra time complexity for noise injection during explanation, and it achieves a competitive time efficiency with C-CHVAE and REVISE. Besides, it is also noted that conducting counterfactual search in latent space (i.e., C-CHVAE, REVISE, DPC) generally has a better time efficiency compared with that in original data space (i.e., DiCE, CADEX), which may due to the fact that latent manifolds are typically in lower dimensions.

\subsection{Evaluation on DPC Protectiveness} 

In this part, we evaluate the protectiveness of DPC on model extraction, membership inference and attribute inference, regarding to the target model $f$ and dataset $\mathcal{D}$. We aim to check whether DPC can effectively relieve the risks on those privacy attacks. 

\subsubsection{Experimental Settings.} 

We assume adversaries purposely collect the derived counterfactuals to conduct a series of privacy attacks, by querying the target model $f$ trained on $\mathcal{D}$. To better evaluate the protectiveness, it is assumed that adversaries have the access to the prediction vectors of $f$ and partially know the data distribution of $\mathcal{D}$. We consider three types of privacy attack~\cite{rigaki2020survey} in experiments, i.e., model extraction, membership inference and attribute inference, where the inference attacks are based on the extracted surrogate model $f^{\prime}$. Without loss of generality, we simply involve DiCE to generate non-DP counterfactuals as references.

\begin{table*}[t]
\small
\renewcommand\arraystretch{1.2}
\centering 
\caption{The accuracy (\%) of the extracted surrogate model under known/unknown-architecture scenarios.} 
\vspace{-0.3cm}
\label{tab:me} 
\begin{tabular}{|p{2.3cm}||cc|cc|cc|cc|cl|cl|}
\hline
\multirow{2}{*}{\textit{Dataset}} & \multicolumn{2}{c|}{\textbf{$|\mathcal{X}_{q}|$=250}} & \multicolumn{2}{c|}{\textbf{$|\mathcal{X}_{q}|$=500}} & \multicolumn{2}{c|}{\textbf{$|\mathcal{X}_{q}|$=1000}} & \multicolumn{2}{c|}{\textbf{$|\mathcal{X}_{q}|$=2000}} & \multicolumn{2}{c|}{\multirow{2}{*}{\textbf{\begin{tabular}[c]{@{}c@{}}Base\\ Extraction\end{tabular}}}} & \multicolumn{2}{c|}{\multirow{2}{*}{\textbf{\begin{tabular}[c]{@{}c@{}}Target\\ Model\end{tabular}}}} \\ \cline{2-9}
                                  & \textit{Non-DP}   & \textit{DPC}  & \textit{Non-DP}   & \textit{DPC}  & \textit{Non-DP}   & \textit{DPC}   & \textit{Non-DP}   & \textit{DPC}   & \multicolumn{2}{c|}{}                                                                                    & \multicolumn{2}{c|}{}                                                                                 \\ \hline
\textbf{Adult}                    & 79.5/78.1         & 73.1/72.3     & 80.1/79.0         & 71.0/70.3     & 81.9/80.2         & 68.9/67.1      & 83.1/81.7         & 66.5/65.1      & \multicolumn{2}{c|}{75.2/73.6}                                                                           & \multicolumn{2}{c|}{85.6}                                                                             \\
\textbf{Hospital}                 & 64.3/63.1         & 61.3/59.8     & 65.7/63.9         & 60.2/59.1     & 66.7/64.8         & 58.9/57.7      & 67.0/65.8         & 56.4/55.2      & \multicolumn{2}{c|}{61.0/59.9}                                                                           & \multicolumn{2}{c|}{70.1}                                                                             \\
\textbf{HomeCredit}               & 73.3/70.9         & 71.0/68.8     & 74.8/72.5         & 70.1/66.9     & 75.6/73.4         & 68.8/65.1      & 76.0/74.8         & 66.5/62.7      & \multicolumn{2}{c|}{70.5/67.7}                                                                           & \multicolumn{2}{c|}{77.9}                                                                             \\
\textbf{MNIST}                    & 84.2/81.6         & 81.0/78.3     & 85.0/83.1         & 80.4/77.3     & 86.5/84.2         & 78.2/74.8      & 87.1/84.9         & 76.4/73.1      & \multicolumn{2}{c|}{81.6/78.1}                                                                           & \multicolumn{2}{c|}{90.8}                                                                             \\
\textbf{Purchase}                 & 56.1/53.6         & 55.4/53.0     & 56.7/54.2         & 55.0/52.7     & 57.3/54.9         & 54.6/52.1      & 58.8/55.3         & 53.8/51.6      & \multicolumn{2}{c|}{55.7/53.1}                                                                           & \multicolumn{2}{c|}{71.3}                                                                             \\
\textbf{Texas}                    & 29.0/25.9         & 28.6/25.8     & 29.3/26.2         & 28.2/25.2     & 29.8/26.9         & 27.7/24.6      & 30.2/28.6         & 27.1/24.1      & \multicolumn{2}{c|}{28.6/25.9}                                                                           & \multicolumn{2}{c|}{51.4}                                                                             \\ \hline
\end{tabular}
\end{table*}

\subsubsection{Model Extraction Attack.} 

The goal of model extraction is to obtain a high-fidelity surrogate model $f^{\prime}$ to approximate the performance of the target model $f$. Counterfactuals can be useful for this type of attack~\cite{aivodji2020model}, since they explicitly reveal abundant information on decision boundaries of $f$. In experiments, we assume adversaries construct their transfer sets for extraction as follows:
\begin{equation}
    \mathcal{S}^{ts} = \left\{\mathcal{X}_{q}, f(\mathcal{X}_{q})\right\} \cup \xi(\mathcal{X}_{q}, f),
\end{equation}
and then use the constructed $\mathcal{S}^{ts}$ to obtain the surrogate $f^{\prime}$ through the pipeline in~\cite{jagielski2020high}. We consider $|\mathcal{X}_{q}| \in \{250, 500, 1000, 2000\}$ for a comprehensive evaluation, and test the accuracy of $f^{\prime}$ under both known-architecture and unknown-architecture scenarios\footnote{The known-architecture scenario indicates $f$ and $f^{\prime}$ use the same neural architecture, while the unknown-architecture scenario does not have such assumption}. 
To simulate the attack scenario, we save some data samples from the whole set for adversaries, and only use part of the data instances for training $f$. Specifically, we merge the original training and test set to construct a big dataset, and then randomly sample $4$ small datasets which are not overlapping between each other. We use one of the small datasets to train/test our $f$, and use the other $3$ sets to conduct attacks. For the known-architecture scenario, we employ the same structure of $f$ for $f^{\prime}$. For the unknown-architecture scenario, we add one more dense layer to $f^{\prime}$ to enforce the difference.
The relevant results are shown in Table~\ref{tab:me}, where the base extraction indicates the attack scenario simply with $\{\mathcal{X}_{q}, f(\mathcal{X}_{q})\}$. In this set of experiments, DPC is evaluated with $\epsilon=0.025$, and Non-DP counterfactuals are derived by DiCE with default settings. Based on the numerical results in Table~\ref{tab:me}, we note that DPC can effectively reduce the accuracy of surrogate $f^{\prime}$ in both known/unknown-architecture scenarios, and such protection becomes stronger as the number of involved DPC samples increases. For cases where the feature space is large (e.g., Purchase and Texas), it is observed that DPC protection may not be that significant compared with the base extraction. The reasons may lie in two folds: (1) Extracting high-fidelity models with large feature space is naturally challenging by itself~\cite{rigaki2020survey}; (2) Protection for large feature space would require an extremely strict budget on DP. Furthermore, it is also noted that non-DP counterfactuals can generally help improve the accuracy of surrogate $f^{\prime}$, which is in line with the observations in~\cite{aivodji2020model}. 

\subsubsection{Membership/Attribute Inference Attack.} 

\begin{figure}[t] 
\centering 
\includegraphics[width=0.9\columnwidth]{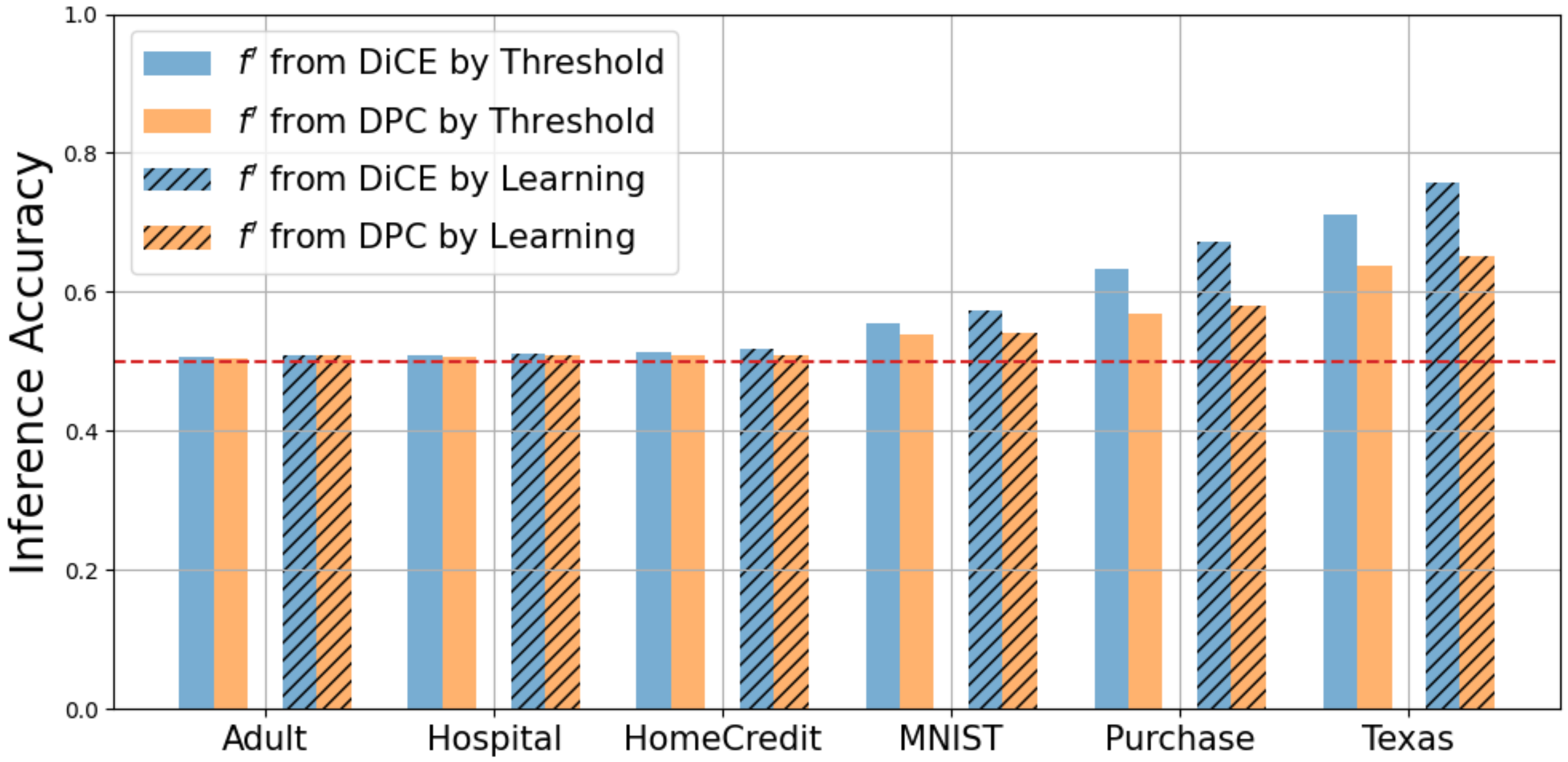} 
\vspace{-0.25cm}
\caption{The accuracy of membership inference with $f^{\prime}$.} 
\label{fig:mia} 
\end{figure}

\begin{figure}[t] 
\centering 
\includegraphics[width=0.9\columnwidth]{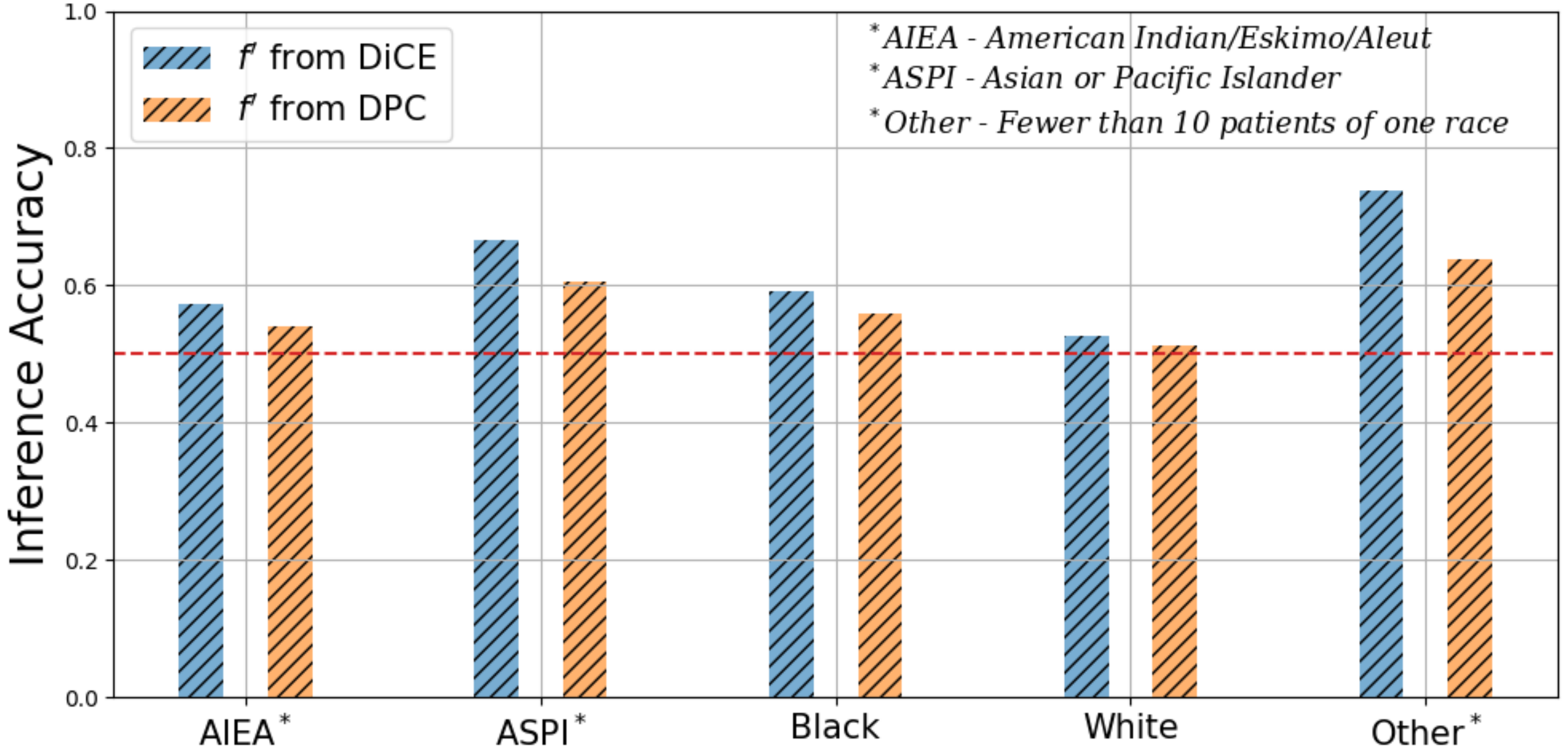} 
\vspace{-0.25cm}
\caption{The attribute inference on \texttt{RACE} in Texas with $f^{\prime}$.} 
\label{fig:aia} 
\end{figure}

\begin{figure}[t] 
\centering 
\includegraphics[width=0.9\columnwidth]{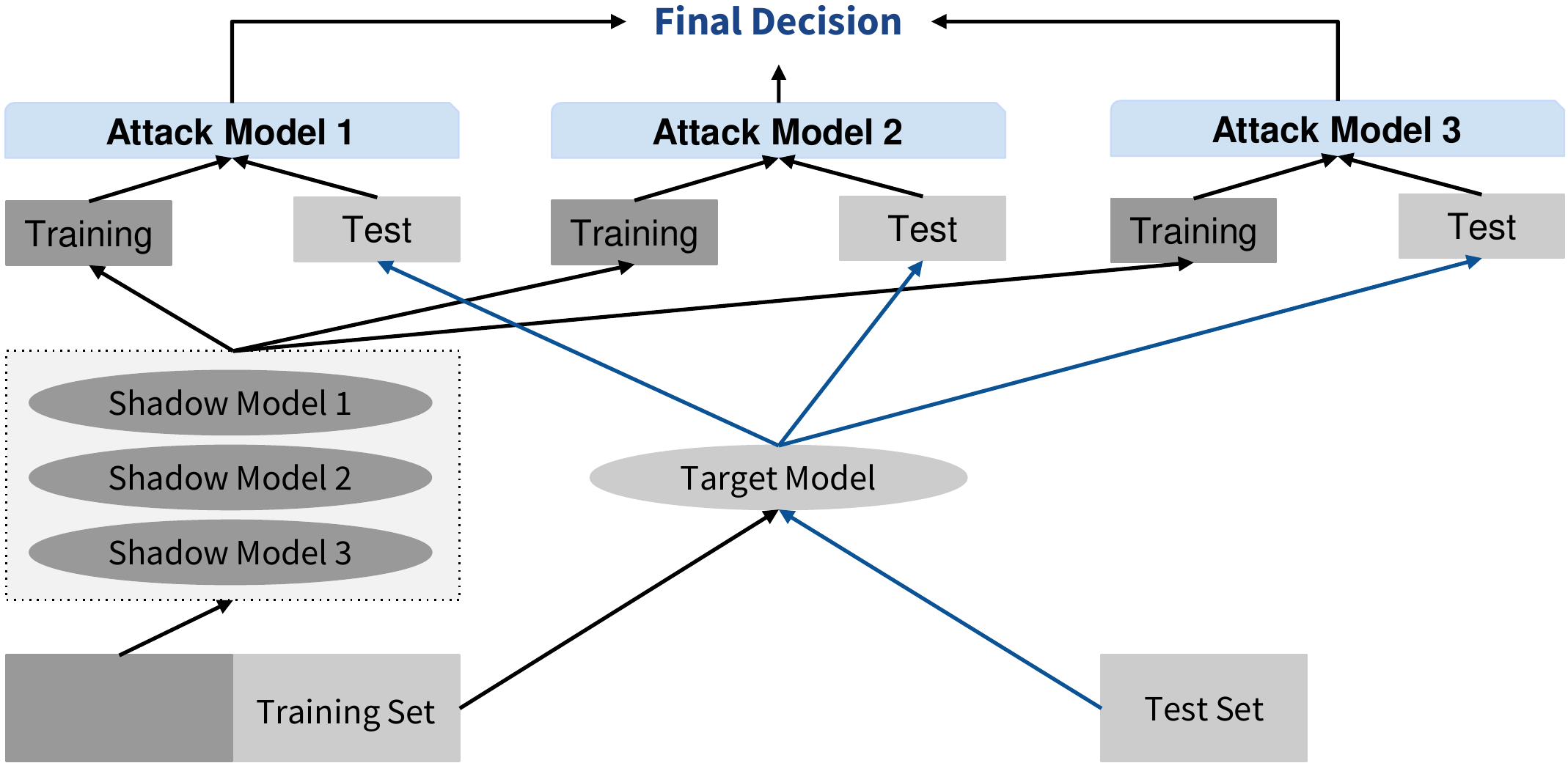} 
\vspace{-0.05cm}
\caption{The learning-based scheme for inference attacks.} 
\label{fig:learnattack} 
\end{figure}

With the extracted surrogate $f^{\prime}$ from counterfactuals, we further conduct experiments on inference attacks to evaluate the DPC protectiveness. We first focus on the membership inference, where the goal is to determine whether a particular instance was used for training of $f$. In experiments, we follow the common settings by creating $4$ small datasets out of the large one, and split each small dataset 50/50 into the training and testing set. In this way, adversaries are assumed to have an a priori knowledge that half of the instances are members of the training set. Specifically, we employ two types of methods for inference, i.e., threshold-based and learning-based algorithms~\cite{rigaki2020survey}, and feed the prediction vector from $f^{\prime}$ as the input. 
For the threshold-based method, our threshold is calculated based on the other $3$ shadow models trained with similar distribution~\cite{shokri2017membership}. For the learning-based method, we follow the scheme shown in Fig.~\ref{fig:learnattack} to conduct attacks. Our attack networks are designed as $[u, 1024, 512, 256, 64, 1]$ in a fully connected manner with ReLU following~\cite{shokri2017membership}, where $u$ indicates the dimension of input sources. We train the attack models for $30$ epochs using Adagrad with a learning rate 1e-2 and a decay rate 1e-7. 
The relevant results are shown in Fig.~\ref{fig:mia}, in which DPC is derived under $\epsilon=0.025$ and used for $f^{\prime}$ extraction with $|\mathcal{X}_{q}|=2000$. From the results, significant protection by DPC can be observed in Purchase and Texas datasets, and the membership inference in other cases almost fail which might be partly because the target models are less overfitted~\cite{yeom2018privacy}. Then, we take the Texas data as a successful case for further studies on attribute inference, and focus on a sensitive attribute `\texttt{RACE}' for evaluation. In this set of attacks, we only employ the learning-based method for inference. The empirical results are shown in Fig.~\ref{fig:aia}. Based on the attack performance, it is observed that DPC can protect the attribute from being recovered by reducing the inference accuracy. Besides, we also note that the minority values (e.g., `\texttt{Other}') are more vulnerable than those majority ones (e.g., `\texttt{White}') for inference, which may due to the sharp regions where the model overfits. The related fairness issues here will be left for our future research.

\subsection{DPC Case Study} 

\begin{figure}[t] 
\centering 
\includegraphics[width=0.8\columnwidth]{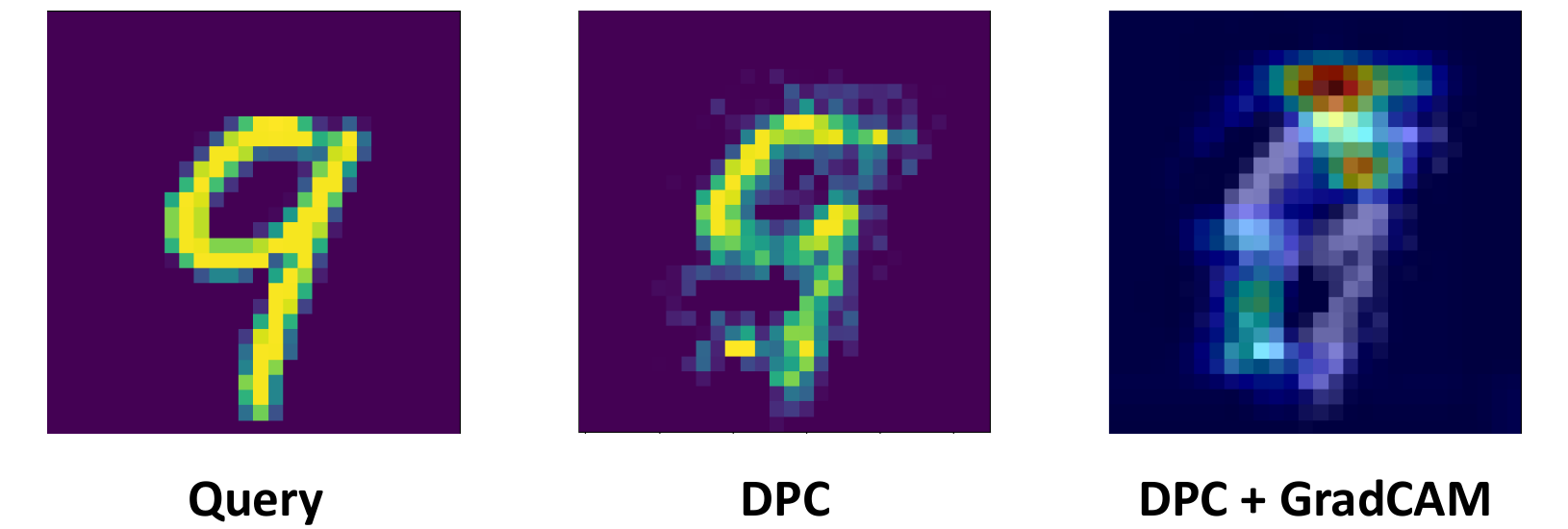} 
\vspace{-0.2cm}
\caption{A case study of DPC for explanation on MNIST.}
\label{fig:dpc_case} 
\end{figure}

We further show a case result of DPC in Fig.~\ref{fig:dpc_case} for explanation on MNIST, aiming to intuitively illustrate how DPC protect the target $f$ and the associated $\mathcal{D}$. As shown in Fig.~\ref{fig:dpc_case}, we counterfactually reason a query instance `\texttt{9}' with the label `\texttt{5}', and our DPC sample is derived with $\epsilon=0.5$. In this case, although the counterfactual label is `\texttt{5}', the predicted label of DPC from $f$ is `\texttt{4}', which successfully blurs the decision boundaries of $f$ for adversaries. To validate the source of protection, we employ the post-hoc method GradCAM~\cite{selvaraju2017grad} to scrutinize the DPC, inspecting which pixels contribute significantly to this misclassification case. From the saliency map, we note highly contributed regions mainly lie in some noisy pixels, demonstrating that the noise injection from the proposed framework plays a key role in such protection. 

\section{Related Work}\label{rw} 

The research related to differentially private model explanation is limited, and only a few work pay attention to this area recently. Overall, those work can be categorized into two groups: (1) Phenomenon studies; (2) Methodology studies. For phenomenon studies, the authors of \cite{milli2019model} first investigates the risks of releasing gradient-based attributions on model reconstruction. Then, such risks from counterfactual explanation are further studied in~\cite{aivodji2020model}. Work~\cite{shokri2021privacy} specifically focuses on the privacy issues raised from the gradient-based~\cite{selvaraju2017grad} and sample-based~\cite{koh2017understanding} explanations. For methodology studies, the authors in~\cite{patel2020model} propose an adaptive DP mechanism to inject noises into the generalized LIME explanation~\cite{ribeiro2016should} by accessing the explanation set, and work~\cite{mochaourab2021robust} proposes to derive explanations with DP by injecting noises to the target model itself. To best of our knowledge, our work is the very first attempt on methodology studies in deriving counterfactuals with DP guarantees.

\section{Conclusions}

In this paper, we have proposed the DPC framework to derive counterfactual explanation with differential privacy guarantees. Specifically, we first construct noisy class prototypes in latent space through an autoencoder trained with the functional mechanism, and then use the prototype to search counterfactual samples for model explanation. Overall, we theoretically prove the DP badge for the proposed framework, and empirically validate the DPC through a set of quantitative and qualitative experiments, showing that DPC can be both \emph{protective} and \emph{informative} as explanations delivered to end-users. Future research originated from DPC may include the extension on other types of data/model interpretation, as well as the related fairness issues in protection.

\newpage
\bibliographystyle{ACM-Reference-Format} 
\bibliography{ref}

\end{document}